\documentclass[runningheads]{llncs}
\usepackage[T1]{fontenc}
\usepackage{graphicx}
\usepackage[table]{xcolor}
\usepackage{colortbl} 
\usepackage{tabularx}
\usepackage{booktabs}
\usepackage{geometry} 
\usepackage{pdflscape}
\usepackage{hyperref}
\usepackage{amsmath} 

\usepackage[normalem]{ulem} 

\begin{document}
\title{\textit{StyloAI}: Distinguishing AI-Generated Content with Stylometric Analysis}

%

\author{Chidimma Opara}
\authorrunning{C. Opara et al.}

\institute{School of Computing, Engineering \& Digital Technologies\\
Teesside University, Middlesbrough, UK\\
\email{c.opara@tees.ac.uk}}

\maketitle              
\begin{abstract}
The emergence of large language models (LLMs) capable of generating realistic texts and images has sparked ethical concerns across various sectors. In response, researchers in academia and industry are actively exploring methods to distinguish AI-generated content from human-authored material. However, a crucial question remains: What are the unique characteristics of AI-generated text? Addressing this gap, this study proposes \textit{StyloAI}, a data-driven model that uses 31 stylometric features to identify AI-generated texts by applying a Random Forest classifier on two multi-domain datasets. \textit{StyloAI} achieves accuracy rates of 81\% and 98\% on the test set of the AuTextification dataset and the Education dataset, respectively. This approach surpasses the performance of existing state-of-the-art models and provides valuable insights into the differences between AI-generated and human-authored texts.

\keywords{Stylometric Features  \and AI in Education \and Natural Language Processing \and ChatGPT.}
\end{abstract}
\section{Introduction}
The use of AI-generated output has sparked a discussion on the ethical implications of AI in various fields, particularly in Education, where traditional assessment methods like quizzes and essays are foundational to student evaluation. However, studies have shown that AI-powered text generation tools can produce essays, reports, or other written assignments with minimal effort \cite{birenbaum2023chatbots}, \cite{khalil2023will}. This raises a pivotal ethical question: Is using AI-generated content in academic settings acceptable?

Recent studies have explored the application of AI in enhancing educational experiences \cite{ngo2023perception}. Alongside this, there has been significant research into distinguishing AI-generated content from human-authored texts. The predominant method for such differentiation utilises deep learning techniques for automatic text classification, involving training models such as CNNs, RNNs, LSTM, and Transformer-based models like Roberta and GPT to categorise text into predefined classes \cite{sarvazyan2023overview}. However, despite their effectiveness, these deep learning approaches are often criticised for their "black box" nature, which conceals the specific linguistic cues indicative of AI authorship. Moreover, the focus of these current state-of-the-art models on specific domains raises concerns about their ability to generalise across different domains.

To address the challenges mentioned above, this study's objectives are twofold: first, to identify stylometric features capable of distinguishing AI-generated content from human-written content, and second, to develop a machine learning model that leverages these identified features to accurately classify texts across various domains as either AI-generated or human-written.

This study makes the following contributions:

\begin{enumerate}
    \item \textit{Identification of stylometric features to Differentiate AI-generated Texts from Human-Authored Content}: This research empirically identified 31 stylometric features across multi-domain datasets. Among these features, 12 represent new contributions to the field which are instrumental in distinguishing between AI-generated and human-authored texts. 

    \item \textit{\textit{StyloAI}: A data-driven model that uses stylometric analysis to attribute authorship of texts within multi-domain datasets}: By applying the identified 31 stylometric features on a Random Forest classifier, \textit{StyloAI}, achieves accuracy rates of 81\% and 98\% on two multi-domain annotated dataset, outperforming existing state-of-the-art models. The study further experimentally validates the significance of \textit{StyloAI}'s features.
\end{enumerate}

The rest of the paper is structured as follows: the next section provides an overview of related works, discussing various techniques proposed for detecting AI-generated texts. Section \ref{methodology} delves into a detailed description of the proposed model. Following that, Section \ref{results} presents a thorough evaluation of the \textit{StyloAI} model. Section \ref{conclusion} concludes the paper.

\section{Related Works}\label{related_works}
Research into distinguishing AI-generated text from human-written content has taken various paths, reflecting a growing interest in developing robust detection tools. One approach involves leveraging industry-developed AI detection tools for text classification. The study by Akram et al. \cite{akram2023empirical} assessed the effectiveness of these tools across diverse datasets, highlighting variations in performance and prompting the need for further innovation in detection methodologies.

Another approach to detecting AI-generated texts is the use of automatic text classification. These studies \cite{fagni2021tweepfake}, \cite{uchendu2020authorship} and \cite{zellers2019defending} applied deep learning models to differentiate between AI-generated and human-authored texts. This method has gained traction for its potential to provide high accuracy rates. However, these are "black box" models and the reliance on domain-specific datasets for training raises questions about their applicability across varied contexts.

A promising yet underexplored approach in distinguishing AI-generated content involves integrating handcrafted linguistic features with shallow machine learning classifiers. Recent studies, such as those by \cite{kumarage2023stylometric} and \cite{mindner2023classification}, have introduced a range of features aimed at differentiating between human-authored and AI-generated texts. Despite the initial promise of these methodologies, researchers continue to strive for a balance between accuracy, interpretability, and scalability in their proposed solutions.

\section{Methodology}\label{methodology} 
\begin{figure}
\includegraphics[width=0.9\textwidth]{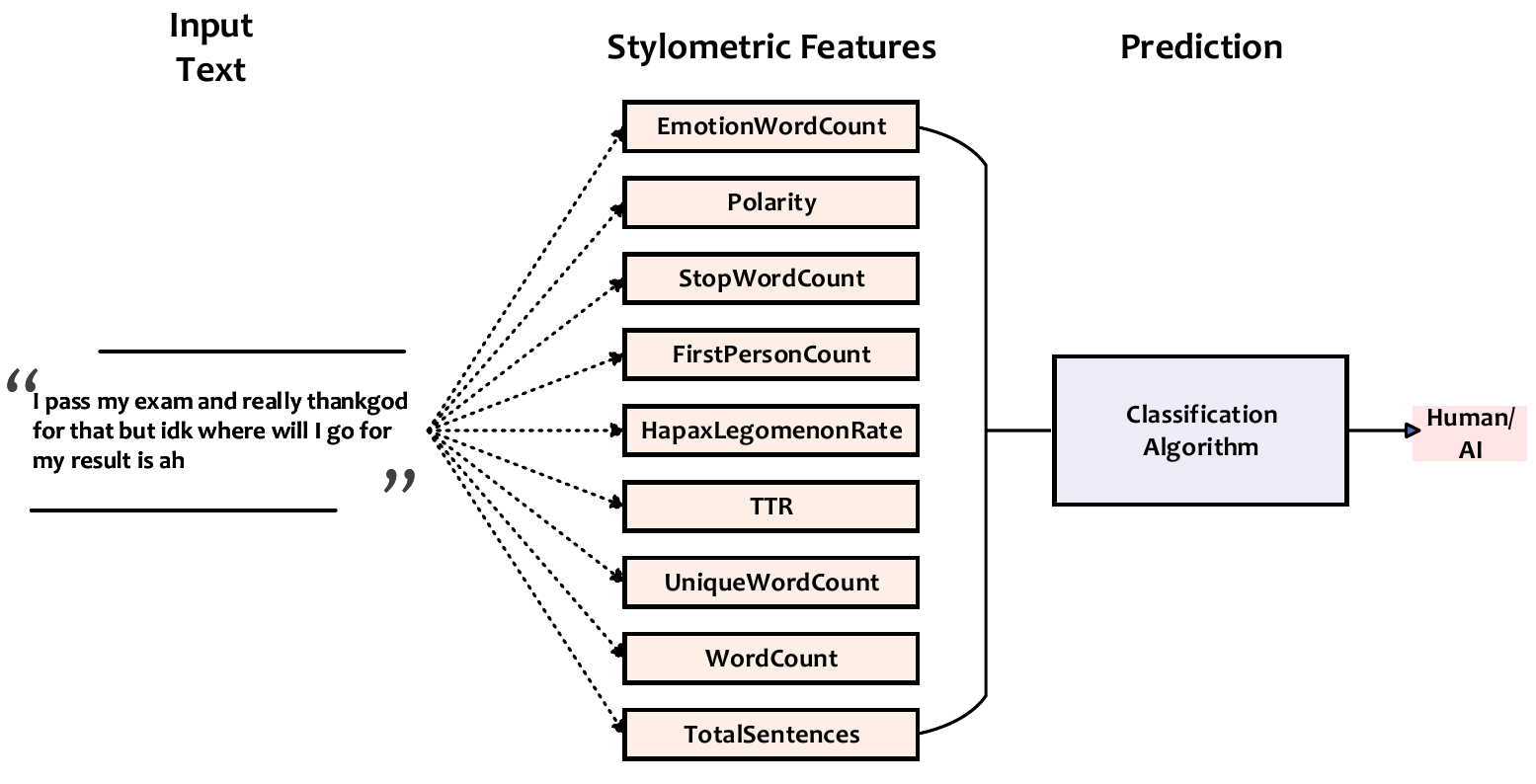}
\caption{The \textit{StyloAI} Process: First, textual characteristics are extracted with a feature-engineering approach. Then, a classification algorithm is used to build the model.} \label{fig:stylo}
\end{figure}

This study aims to design and develop a technique to identify AI-generated texts. Stylometric features are extracted from texts to make predictions about their authorship. This section first discusses the feature set used in \textit{StyloAI}, followed by the machine learning algorithms considered. It concludes with the datasets employed to evaluate the proposed model. Figure \ref{fig:stylo} illustrates the process within the \textit{StyloAI} model.

\subsection{\textit{StyloAI} feature Set}

The feature set within the \textit{StyloAI} model comprises 31 stylometric features, 12 of which are new to the detection of AI-generated texts. These features are categorised into six main groups for clarity and ease of understanding. Table \ref{tab:combined_features} summarises these features.

6 features categorised under \textbf{Lexical Diversity} focus on the range and distribution of words used within a text, highlighting differences in vocabulary richness between AI-generated and human-written content. This study introduces the \textit{Type-Token Ratio} (TTR) and \textit{Hapax Legomenon Rate }(HLR) to evaluate lexical diversity in detecting AI-generated texts.  Additionally, basic metrics obtained from previous studies \cite{kumarage2023stylometric}, \cite{mindner2023classification} such as \textit{WordCount}, \textit{UniqueWordCount}, and \textit{CharacterCount}, assess the total number of words, unique words, and characters present in a text, providing further insights into its lexical complexity.

\textbf{Syntactic Complexity} with 12 features, assesses the structural aspects of language, such as sentence length and complexity, to identify patterns unique to human or AI authors.
Some features included in this category are the \textit{ComplexVerbsCount} and \textit{ContractionCount}. The hypothesis behind these features is that capturing verbs not in the first most common verbs, using multiple commas, and conjunctions like "and" or "but" indicate more structurally intricate sentences characteristic of human written texts. 

The 4 features within the \textbf{Sentiment and Subjectivity} category analyse the emotional tone and subjective expressions within the text, providing insights into how emotions and opinions are conveyed. The \textit{Polarity} and \textit{Subjectivity} were obtained from previous studies \cite{kumarage2023stylometric}, \cite{mindner2023classification} while the \textit{VADERCompound} and the count of words (\textit{EmotionWordCount}) that reflect emotions, including "fear", "joy", and "sadness" were introduced in this study. The hypothesis behind these features is that AI-generated texts might not match the emotional depth of human-written texts. Also, anomalies in emotional word usage, such as overuse, underuse, or misuse in context, could hint at non-human authorship.

The \textbf{Readability} comprising 2 features, evaluates how easy it is to understand the text by considering the intended audience's complexity level and educational background. Scores that have been used in other studies \cite{mindner2023classification}, such as the \textit{FleschReadingEase} and the metric introduced in this study, the \textit{GunningFogIndex}, are based on factors such as sentence length and word complexity. These metrics provide quantifiable assessments of text accessibility for readers with diverse levels of education and literacy. The differences in the intended audience or content sophistication can be identified by comparing readability scores between human and AI-generated texts.
The \textbf{Named Entities} category with 4 features focuses on the presence and usage of specific names, places, and organisations, which can vary significantly between AI-generated and human-written texts. This study introduced \textit{DateEntities} and \textit{DirectAddressCount} features. The hypothesis is that incorrect, anachronistic, or overly vague references to dates can be a red flag. AI may either fail to correctly anchor events in time or use dates in ways that don't align with known historical or contextual facts. Also, human-written text should have many instances where the text directly addresses the reader or another hypothetical listener.

Finally, the 3 features grouped under the \textbf{Uniqueness and Variety} category focus on how original and diverse the content is, thereby distinguishing between repetitive AI-generated patterns and the dynamic variability seen in human writing. This study introduced the uniqueness of the bigrams/trigrams feature, quantifying how frequently new or varied word pairings appear in the text. Texts with a higher bigram/trigram uniqueness ratio may be considered more creative or novel, as they employ a more comprehensive range of lexical combinations, as expected in human-written texts.

\begin{table}[!ht]
\centering
\caption{Comprehensive Features of \textit{StyloAI} Across Six Categories}
\label{tab:combined_features}
\scriptsize 
\begin{tabularx}{\textwidth}{l|l|X|l} 
\hline 
\textbf{Category} & \textbf{Features} & \textbf{Description} & \textbf{Reference} \\
\hline
Lexical & \textit{WordCount} & The total number of words in the text. & \cite{mindner2023classification}\\
& \textit{UniqueWordCount} & The number of unique words used in the text. & \cite{mindner2023classification} \\
& \textit{CharCount} & The total number of characters in the text, including spaces and punctuation. & \cite{mindner2023classification} \\
& \textit{AvgWordLength} & Calculated as $\frac{\text{CharacterCount}}{\text{WordCount}}$. & \cite{mindner2023classification} \\
& \textit{TTR} & Measures lexical diversity, calculated as $\text{TTR} = \frac{\text{UniqueWordCount}}{\text{WordCount}}$. & \textbf{new}\\
& \textit{HapaxLegomenonRate} & The proportion of words that appear only once in the text, $\frac{\text{Number of Words Appearing Once}}{\text{Total Words}}$. & \textbf{new}\\
\hline
Syntactic & \textit{SentenceCount} & The total number of sentences in the text. \\
& \textit{AvgSentenceLength} & Calculated as $\frac{\text{WordCount}}{\text{SentenceCount}}$. & \cite{mindner2023classification} \\
& \textit{PunctuationCount} & The total number of punctuation marks in the text. & \cite{mindner2023classification} \\
& \textit{StopWordCount} & The total number of commonly used words. & \cite{mindner2023classification} \\
& \textit{AbstractNounCount} & The number of nouns representing intangible concepts or ideas. & \textbf{new} \\
& \textit{ComplexVerbCount} & The number of verbs not in the most common 5000 words. & \textbf{new} \\
& \textit{SophisticatedAdjectiveCount} & The number of adjectives with complex suffixes like "ive", "ous", "ic". & \textbf{new}\\
& \textit{AdverbCount} & The total number of adverbs in the text. & \cite{mindner2023classification}\\
& \textit{ComplexSentenceCount} & The number of sentences with more than one clause, indicating complex sentence structures. & \cite{kumarage2023stylometric}\\
& \textit{QuestionCount} & The total number of questions, as indicated by question marks in the text. & \cite{mindner2023classification}, \cite{kumarage2023stylometric}\\
& \textit{ExclamationCount} & The total number of exclamations, as indicated by exclamation marks in the text. & \cite{mindner2023classification}, \cite{kumarage2023stylometric} \\
& \textit{ContractionCount} & The total number of contractions in the text, such as "don't" and "can"t". & \textbf{new} \\
\hline
Sentiment & \textit{EmotionWordCount} & The total number of words associated with emotions in the text. & \textbf{new} \\
& \textit{Polarity} & Measures the text's sentiment orientation (positive, negative, or neutral). & \cite{mindner2023classification}, \cite{kumarage2023stylometric}\\
& \textit{Subjectivity} & Measures the amount of personal opinion and factual information in the text. & \cite{mindner2023classification}, \cite{kumarage2023stylometric} \\
& \textit{VaderCompound} & A sentiment analysis score that combines the positive, negative, and neutral scores to give a single compound sentiment score. & \textbf{new}\\
\hline
Readability & \textit{FleschReadingEase} & Calculated as \( 206.835 - 1.015 \left( \frac{\text{Total Words}}{\text{Total Sentences}} \right) - 84.6 \left( \frac{\text{Total Syllables}}{\text{Total Words}} \right) \). & \cite{mindner2023classification} \\
& \textit{GunningFog} & Estimates the years of formal education needed to understand a text on the first reading, calculated as \( 0.4 \left( \frac{\text{Word Count}}{\text{Sentence Count}} + 100 \left( \frac{\text{Complex Words Count}}{\text{Word Count}} \right) \right) \). & \textbf{new} \\
\hline
Named Entity & \textit{FirstPersonCount} & The number of first-person pronouns. & \cite{mindner2023classification}, \cite{kumarage2023stylometric}  \\
& \textit{DirectAddressCount} & The number of instances where the text directly addresses the reader or another hypothetical listener. & \textbf{new}\\
& \textit{PersonEntities} & The count of named individuals mentioned in the text. & \cite{kumarage2023stylometric}\\
& \textit{DateEntities} & The count of date references within the text. & \textbf{new}\\
\hline
Uniqueness & \textit{Bigram/trigramUniquenes}s  & These measures calculate the uniqueness of two-word and three-word combinations, indicating the originality and creative combinations of words in the text. & \textbf{new} \\
& \textit{SyntaxVariety }& The count of all the POS tags in a text. &\cite{kumarage2023stylometric}\\
\hline
\end{tabularx}
\end{table}

\subsection{Model Selection and Implementation}

The task of detecting AI-generated text was approached as a binary classification problem, where known AI-generated texts were treated as negative samples and human-written texts as positive samples. Several popular binary classification techniques in machine learning, focusing on these popular options, were explored: Random Forest, SVM, Logistics Regression, Decision Tree, KNN classification and Gradient Boosting.

All experiments for \textit{StyloAI} were conducted on the Google Colab platform, using machine learning classifiers with default settings from the Scikit-Learn library\footnote{\url{https://scikit-learn.org/stable/}}. For reliable outcomes, a 5-fold cross-validation method was employed, dividing our dataset in each iteration into 80\% for training, 10\% for validation to fine-tune hyperparameters, and 10\% for testing with unseen texts.

\subsection{Datasets}
This study employed two datasets to analyse and evaluate the proposed model. The first, termed the AuTexTification dataset, is from the study by Sarvazyan et al.'s \cite{sarvazyan2023overview}, while the second dataset is drawn from Mindner et al.'s study \cite{mindner2023classification}. 

The Subtask 1 AuTexTification dataset comprises texts from humans and large language models across five domains: tweets, reviews, how-to articles, news, and legal documents sourced from publicly available datasets like MultiEURLEX, XSUM, and Amazon Reviews, among others. This dataset, detailed further in the published article and its \href{https://github.com/autextification/AuTexTification-Overview/tree/main/datasets}{official repository}, comprises 27,989 AI-generated and 27,688 human-written texts, totalling \textbf{55,677} entries, making it one of the largest and most varied corpora used for distinguishing between AI-generated and human-authored content.

The second dataset utilised for assessing \textit{StyloAI} was a smaller, multidomain set curated by Mindner et al. \cite{mindner2023classification}, chosen for its relevance to the education sector. This dataset aims to distinguish between human-crafted and AI-generated texts in an educational context. Covering a diverse array of subjects such as biology, chemistry, geography, history, IT, music, politics, religion, sports, and visual arts, it consists of 100 AI-generated texts collected from ChatGPT 3.5 \footnote{\url{https://chat.openai.com/chat}} interactions and 100 human-written texts collected from Wikipedia\footnote{\url{https://en.wikipedia.org/wiki/Main_Page}}, resulting in a well-balanced dataset of \textbf{200} entries. This dataset makes it particularly useful for evaluating \textit{StyloAI}'s performance in educational settings. Further details on this dataset can be found in the original publication and its \href{https://github.com/LorenzM97/human-AI-generatedTextCorpus}{official repository}.

\section{Results}\label{results}
In line with previous studies, The performance of \textit{StyloAI} model was accessed using key metrics, including accuracy, recall, precision, and F1-score. 

Tables \ref{tab:performance_metrics} presents the performance of the extracted features on different shallow learning models on the Subtask 1 AuTexTification dataset \cite{sarvazyan2023overview}. From Table \ref{tab:performance_metrics}, it is evident that the Random Forest classifier consistently outperforms other shallow machine learning algorithms, achieving an 81\% value across all metrics. Conversely, the Logistic Regression model demonstrates the lowest performance, scoring 71\% across all metrics.

\textbf{Note:} From the analysis above, the extracted stylometric features showed optimal performance with the Random Forest classifier. For the rest of this paper, whenever the \textit{StyloAI} model is used, it indicates the model trained with the Random Forest Classifier.

\begin{table}[ht]
\centering
\caption{Performance of Stylometric Features on Selected Shallow Machine Learning Classifiers Using Sarvazyan et al.'s \cite{sarvazyan2023overview} AuTexTification Dataset}
\label{tab:performance_metrics}
\footnotesize
\begin{tabular}{l|c|c|c|c|c}
\hline
\textbf{Model} & \textbf{Precision} & \textbf{Recall} & \textbf{F1-Score} & \textbf{Accuracy} & \textbf{AUC Score} \\
\hline
\textbf{Random Forest} & \textbf{0.81} & \textbf{0.81} & \textbf{0.81} & \textbf{0.81} & \textbf{0.88} \\
SVM & 0.79 & 0.79 & 0.79 & 0.79 & 0.87 \\
Logistic Regression & 0.71 & 0.71 & 0.71 & 0.71 & 0.77 \\
Decision Tree & 0.72 & 0.72 & 0.72 & 0.71 & 0.85 \\
KNN Classification & 0.73 & 0.73 & 0.73 & 0.73 & 0.80 \\
Gradient Boosting & 0.78 & 0.77 & 0.77 & 0.85 & 0.71 \\
\bottomrule
\end{tabular}
\end{table}

Table \ref{tab:baseline_comparison} presents a comparative analysis of \textit{StyloAI} against state-of-the-art using the Subtask 1 AuTexTification dataset. The comparison includes the TALN-UPF model, which integrates probabilistic token-level features from various GPT-2 models, linguistic features like word frequencies and grammar errors, alongside text representations from pre-trained encoders. Also evaluated are approaches such as random baselines, zero-shot (SBZS) and few-shot (SB-FS) based on text and label embedding similarities, bag-of-words with logistic regression (BOW+LR), Low Dimensional Semantic Embeddings (LDSE), and finely tuned language-specific transformers. A deeper exploration of these baseline models can be found in Sarvazyan et al.'s \cite{sarvazyan2023overview} study.

\begin{table}[ht]
\centering
\begin{minipage}[b]{0.45\linewidth} 
\centering
\caption{Comparison of \textit{StyloAI} with Sarvazyan et. al's \cite{sarvazyan2023overview} Models on the AuTexTification Dataset}
\label{tab:baseline_comparison}
\footnotesize
\begin{tabular}{l|lc}
\hline
\textbf{Model} & \textbf{F1-Score} \\
\hline
\textbf{\textit{StyloAI}} & \textbf{0.81}\\
 TALN-UPF & 0.80\\
 BOW+LR & 0.74 \\
 LDSE & 0.60 \\
 SB-FS & 0.59 \\
 Transformer & 0.57 \\
 Random & 0.50 \\
 SB-ZS & 0.33 \\
\bottomrule
\end{tabular}
\end{minipage}
\hfill
\begin{minipage}[b]{0.45\linewidth} 
\centering
\caption{Comparison of \textit{StyloAI} with Mindner et. al's \cite{mindner2023classification} Models on the Education Dataset}
\label{tab:performance_metrics_midner}
\footnotesize
\begin{tabular}{l|c|c}
\hline
\textbf{Model} & \textbf{F1-Score} & \textbf{Accuracy} \\
\hline
\textbf{\textit{StyloAI}} & \textbf{0.97} & \textbf{0.98} \\
Mindner et. al's Features + XGBoost & 0.90 & 0.90 \\
Mindner et. al's Features + Random Forest & 0.98 & 0.98 \\
Mindner et. al's Features + MLP & 0.87 & 0.87 \\
\bottomrule
\end{tabular}
\end{minipage}
\end{table}

Among the compared models, TALNUP performed with an F1-score of 0.80. Notably, TALNUP relies on deep learning and word vectors. However, as shown in Table \ref{tab:baseline_comparison}, \textit{StyloAI} surpasses it with an F1-score of 0.81. Moreover, StyloAI significantly outperforms the least effective baseline model—the zero-shot model—by 48\% in the F1 score, highlighting its superior ability to distinguish AI-generated texts from human-written content.

In the evaluation of \textit{StyloAI} on the second dataset created by Mindner et al., Table \ref{tab:performance_metrics_midner} illustrates that \textit{StyloAI} performs comparably to the model proposed by Mindner et al. \cite{mindner2023classification}, despite Mindner et al.'s reliance on AI-generated features. Achieving a 98\% on accuracy demonstrates the robustness of \textit{StyloAI } when applied to text generated from the education sector. Furthermore, in their study, Mindner et al. compared their model's performance with an industry-developed AI detector, revealing that their model outperforms GPTZero by 28.9\% relatively in accuracy and 24.2\% relatively in F1-score. The hypothesis is that \textit{StyloAI} will have similar performance when compared with GPTZero since its performance is comparable with the result of Mindner et al. on the base dataset.

\subsubsection{Feature Importance}
\begin{figure}
\includegraphics[width=0.8\textwidth]{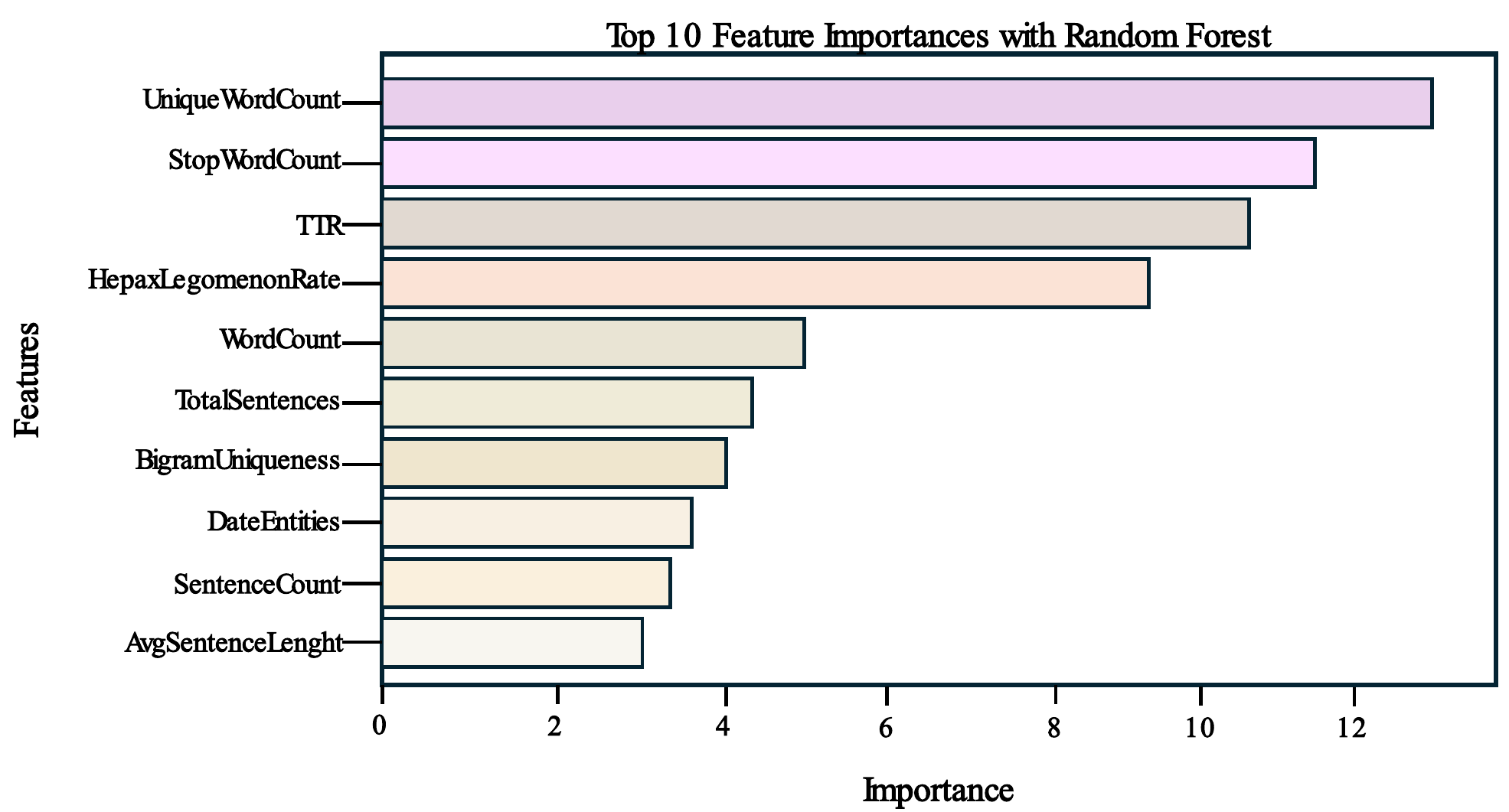}
\caption{Ranking of Features and How they Influence the Performance of the Random Forest Classifier} \label{fig:importance}
\end{figure}

From Figure \ref{fig:importance}, it is observed that the 4 most significant features highlighted by the Random Forest algorithm are the \textit{UniqueWordCount}, \textit{StopWordCount}, \textit{TTR}, and \textit{HapaxLegomenonRate}. 

These top 4 features are pivotal in distinguishing AI-generated texts from human-authored ones. The \textit{UniqueWordCount}, which counts non-repetitive words, can reveal AI's tendency to use rare words excessively, aiming for sophistication but resulting in an unnatural count. In contrast, humans write sentences with more stop words to maintain a natural flow. While AI models strive to replicate this pattern, they may deviate due to a lack of human speech rhythm or misuse of stop words.

\textit{TTR}, which measures lexical diversity, shows that human texts achieve a balance in vocabulary that suits the text's length and complexity. In contrast, AI-generated texts often miss this balance by excessively expanding or restricting their vocabulary range. Additionally, the \textit{HapaxLegomenonRate}, indicating the use of words appearing only once, signifies rich and detailed vocabulary in human writing—a trait that AI texts may not consistently emulate, thereby impacting their authenticity.

\section{Conclusion}\label{conclusion}
This paper explored stylometric features to classify text as AI-generated or human-written. Specifically, 31 features categorised under Lexical Diversity, Sentiment and Subjectivity, Readability, Named Entities and Uniqueness and Variety were extracted to determine what characterises AI-generated texts. The proposed model, \textit{StyloAI}, comprising the 31 features trained on a random forest classifier, provided an accuracy of 81\% and 98\% on two multi-domain datasets. This study also demonstrates that a balanced use of non-repetitive words and a vocabulary that suits the text length and complexity are the top features that differentiate AI-generated texts from human-authored ones. 
A limitation of the study is that it uses the AuTexTification dataset, generated using the Bloom model rather than the more commonly used GPT models (GPT-3.5 and GPT-4). Future work will focus on investigating features from datasets generated by GPT models.

\bibliographystyle{splncs04}
\bibliography{AIED}

\begin{thebibliography}{10}
\providecommand{\url}[1]{\texttt{#1}}
\providecommand{\urlprefix}{URL }
\providecommand{\doi}[1]{https://doi.org/#1}

\bibitem{akram2023empirical}
Akram, A.: An empirical study of ai generated text detection tools. arXiv preprint arXiv:2310.01423  (2023)

\bibitem{birenbaum2023chatbots}
Birenbaum, M.: The chatbots’ challenge to education: Disruption or destruction? Education Sciences  \textbf{13}(7), ~711 (2023)

\bibitem{fagni2021tweepfake}
Fagni, T., Falchi, F., Gambini, M., Martella, A., Tesconi, M.: Tweepfake: About detecting deepfake tweets. Plos one  \textbf{16}(5),  e0251415 (2021)

\bibitem{khalil2023will}
Khalil, M., Er, E.: Will chatgpt g et you caught? rethinking of plagiarism detection. In: International Conference on Human-Computer Interaction. pp. 475--487. Springer (2023)

\bibitem{kumarage2023stylometric}
Kumarage, T., Garland, J., Bhattacharjee, A., Trapeznikov, K., Ruston, S., Liu, H.: Stylometric detection of ai-generated text in twitter timelines. arXiv preprint arXiv:2303.03697  (2023)

\bibitem{mindner2023classification}
Mindner, L., Schlippe, T., Schaaff, K.: Classification of human-and ai-generated texts: Investigating features for chatgpt. In: International Conference on Artificial Intelligence in Education Technology. pp. 152--170. Springer (2023)

\bibitem{ngo2023perception}
Ngo, T.T.A.: The perception by university students of the use of chatgpt in education. International Journal of Emerging Technologies in Learning (Online)  \textbf{18}(17), ~4 (2023)

\bibitem{sarvazyan2023overview}
Sarvazyan, A.M., Gonz{\'a}lez, J.{\'A}., Franco-Salvador, M., Rangel, F., Chulvi, B., Rosso, P.: Overview of autextification at iberlef 2023: Detection and attribution of machine-generated text in multiple domains. arXiv preprint arXiv:2309.11285  (2023)

\bibitem{uchendu2020authorship}
Uchendu, A., Le, T., Shu, K., Lee, D.: Authorship attribution for neural text generation. In: Proceedings of the 2020 Conference on Empirical Methods in Natural Language Processing (EMNLP). pp. 8384--8395 (2020)

\bibitem{zellers2019defending}
Zellers, R., Holtzman, A., Rashkin, H., Bisk, Y., Farhadi, A., Roesner, F., Choi, Y.: Defending against neural fake news. Advances in neural information processing systems  \textbf{32} (2019)

\end{thebibliography}

\end{document}